# Prompt-to-Parts:
# Generative AI for Physical Assembly and Scalable Instructions

David Noever
PeopleTec, 4901-D Corporate Drive, Huntsville, AL, USA, 35805
david.noever@peopletec.com

**ABSTRACT**

We present a framework for generating physically realizable assembly instructions from natural language descriptions. Unlike unconstrained text-to-3D approaches, our method operates within a discrete parts vocabulary, enforcing geometric validity, connection constraints, and buildability ordering. Using LDraw as a text-rich intermediate representation, we demonstrate that large language models can be guided with tools to produce valid step-by-step construction sequences and assembly instructions for brick-based prototypes of more than 3000 assembly parts. We introduce a Python library for programmatic model generation and evaluate buildable outputs on complex satellites, aircraft, and architectural domains. The approach aims for demonstrable scalability, modularity, and fidelity that bridges the gap between semantic design intent and manufacturable output. Physical prototyping follows from natural language specifications. The work proposes a novel elemental *lingua franca* as a key missing piece from the previous pixel-based diffusion methods or computer-aided design (CAD) models that fail to support complex assembly instructions or component exchange. Across four original designs, this novel "bag of bricks" method thus functions as a physical API: a constrained vocabulary connecting precisely oriented brick locations to a "bag of words" through which arbitrary functional requirements compile into material reality. Given such a consistent and repeatable AI representation opens new design options while guiding natural language implementations in manufacturing and engineering prototyping.

## 1. INTRODUCTION

The conversion of natural language into functional physical prototypes represents an emerging frontier in computational design. While text-to-image and text-to-3D systems have advanced significantly [1-14], the challenge of generating physically buildable, materially constrained structures remains open [15-25]. LEGO bricks [1,26-32], with their standardized geometry, tolerances, and global availability, offer a uniquely well-bounded medium for exploring this problem.

LEGO systems are already widely adopted in engineering education [30-31], robotics experiments [19,33-34], and low-cost scientific instrumentation [35-36]. The largest LEGO set with instructions (Eiffel Tower) totals around 10,000 bricks, 58.5-inch height, and nearly 1000 assembly steps (or manual pages). As illustrated in Figure 1, these composable examples highlight the potential for LEGO to serve as an accessible platform for rapid, low-cost experimentation in mechanical design, instrumentation, and physical sciences. The underlying insight is that discretized construction systems—whether LEGO, modular satellites, or flat-pack furniture—share a common structure amenable to language-driven synthesis: a finite vocabulary of parts, a grammar of valid connections, and functional constraints that map to inventive principles (Figure 2).

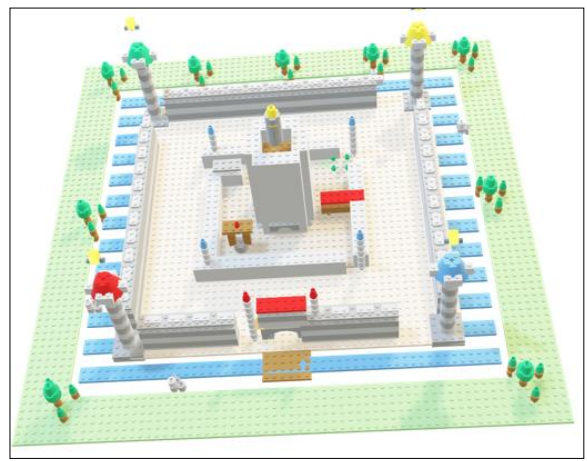

*Figure 1. Complex medieval castle automatically generated as an 860-part instruction kit and bill of materials*

A key question arises: *Can large language models (LLMs) reliably generate accurate LEGO models and step-by-step assembly instructions from arbitrary text prompts?* If successful, this capability would operationalize a form of text-to-prototype, allowing designers, students, and researchers to translate conceptual descriptions directly into testable physical artifacts. The problem is nontrivial. It requires spatial reasoning [17-20], long-horizon planning [34], constraint satisfaction [10,14], and an implicit understanding of real-world forces [33-35] and geometric compatibility [1,3,10]. It also requires the model to produce instructions that a human can follow and that result in a stable assembly [23,29] with no ambiguous or impossible steps.

We hypothesize that *the adoption of a compact, human-readable component language can substantially improve the reliability and scale of LLM-generated physical assemblies.* The hypothesis draws from LLM methods to encode chess games with Forsyth-Edwards Notation (FEN) [37] or use natural language to produce structured query languages (SQL) in databases [38])

Just as FEN encodes complete board states in a single line of text that both humans and machines can parse unambiguously, the LDraw format [39] encodes LEGO assemblies through standardized part identifiers, precise coordinates, and rotation matrices in a syntax that predates and is independent of any particular AI system.

We extend the success of the natural language choice—so called "Bag of Words (BOW)" approaches-- to solve syntax constraints and the need for a similar "Bag of Bricks" method in the physical assembly of parts.

This intermediate representation offers three critical properties absent from unconstrained natural language or mesh-based 3D formats: (1) discrete parts from a finite, well-documented vocabulary where each element has known dimensions and connection geometry; (2) explicit spatial coordinates that eliminate ambiguity about placement; and (3) sequential structure that maps directly to build order.

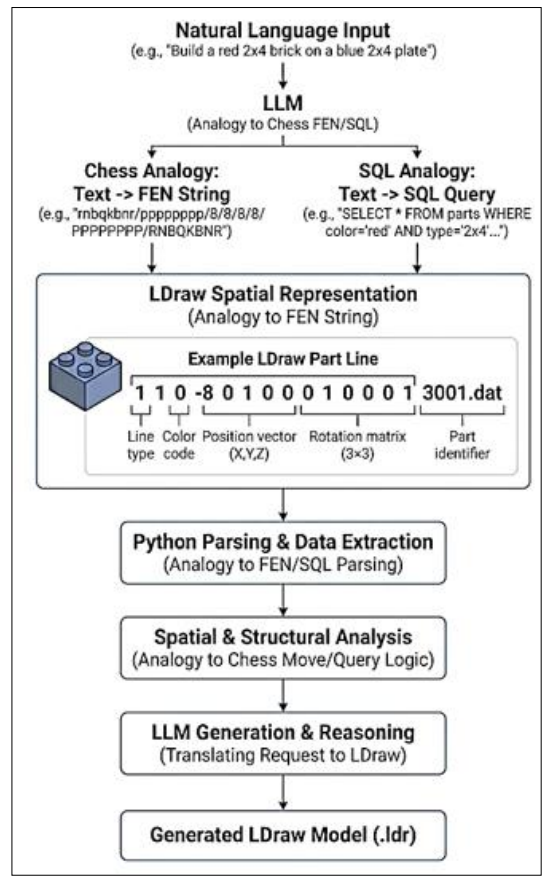

*Figure 2. LLM-Driven LEGO Construction: From Natural Language to LDraw Spatial Reasoning*

We propose that by targeting LDraw as a compilation target—rather than generating prose instructions or continuous 3D representations—LLMs can leverage the format's inherent constraints to produce assemblies of practical complexity, extending beyond the 50–100-part demonstrations typical of current systems toward builds *exceeding 1,000 components with verifiable structural validity (Figure 1).*

To test this hypothesis, we develop a benchmark framework that evaluates LLM performance across three orthogonal dimensions: drawing accuracy (syntactic correctness of LDraw output), structural model validity (physical stability, connectivity, and constraint satisfaction), and instructional coherence (whether the generated build sequence is complete, unambiguous, and executable by a human builder).

We introduce a "bag-of-bricks" partial-credit methodology inspired by bag-of-words evaluation in natural language processing, which isolates modular spatial reasoning competencies by constraining models to fixed part inventories and scoring local correctness independent of global structure. Additionally, we incorporate TRIZ-inspired modification tasks [35] that evaluate whether LLMs can improve existing designs according to engineering tradeoffs—segmentation, dynamization, counterforce, and curvilinear transformation—positioning AI-generated LEGO prototypes not merely as static models but as testable mechanisms for exploring structural and mechanical principles.

Table 1 contrasts prior work across text-to-3D generation, physics-aware synthesis, spatial reasoning benchmarks, and assembly instruction generation, identifying the specific gaps that motivate our approach and evaluation framework (Appendix C)

*Table 1 Gaps in prior work and contributions of the present study.*

| Dimension | Prior Work | Gap | Present Work |
|---|---|---|---|
| Realizability | Text-to-3D/CAD achieves visual plausibility | No physical buildability in discrete modular systems | LDraw as buildable IR with standardized parts and coordinates |
| Stability | Physics-aware generation achieves 98.8% stability [14] | Restricted part vocabulary; no assembly sequencing | Five shape families with step-by-step build orders |
| Spatial Reasoning | Planning accuracy → 0% as sequence length increases [15] | No diagnostic methodology for reasoning competencies | Bag-of-bricks evaluation isolates modular reasoning |
| Instructions | Optimizes for robotic execution success | Human followability not evaluated | Instructional coherence (I-score) as first-class metric |
| Benchmarks | Evaluates generation quality or manipulation separately | No benchmark for constrained buildable assembly generation | Three-axis framework (D/M/I-scores) + TRIZ modification tasks |

## 2. RESULTS AND DISCUSSION

The approach for designing new assemblies with instructions is three steps: 1) formulate a prompt as text or alternatively an existing 2d image that a foundational model (like Anthropic Claude Opus 4.5, 2025) can describe; 2) convert the prompt using our custom python library to configure build steps in LDraw notation; 3) output a .ldr file and analyze its ordered-steps using Leo CAD, Blender, or other open source rendering tools.

To illustrate the complete LLM tool-enabled representation for design and instruction, Figures 1 and 3 show a prompt-created medieval castle design as a simple output of 860 parts with a complete 83 step assembly instruction. Each step represents a standard LEGO page with multiple (5-10 part) assemblies completed in correct order. The design mixes the typical bag of bricks with color, shape, and non-symmetric patterns that add to its overall complexity in three-dimensions.

## 2.1. Prompt-to-Part at Scale.

Appendix A and Figure 4 highlight the scalability of parts assembly, with International Space Station (ISS). The input prompt was "*construct the large-scale detailed ISS model in LEGO*" which in turn used the python library to generate a 3,122-part kit with 112-page instruction manual as detailed build steps and rendered LDraw (Leo CAD) and OBJ Wavefront CAD models (Blender) shown in Figure 4.

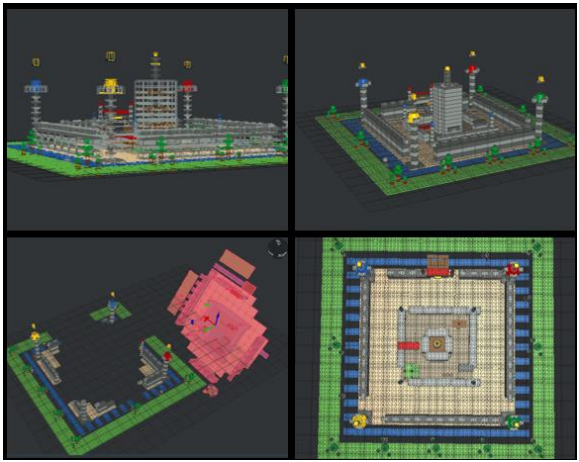

*Figure 3. Detachable build modules and part isolation for physical testing and reassembly*

Appendix A details each national ISS module and its bill of materials in parts illustrate the ordered assembly. Video construction as well as print assets were constructed to assist in final construction from the natural language prompt. The case study highlights the scale of thousands of parts, hundreds of steps, and ordering instructions. The constraint of modular menus of components makes the final outcome consistent as "bag of bricks" derived from "bag of words". This consistency differs in approach from previous methods to align natural language to pixels (VLMs) as tokens or pattern recognition (fine-tuning on LDraw instructions). The intermediate assistance of a python tool provides a library of available and legal moves.

## 2.2. Prompt-to-Part as Modular Modifications.

Appendix B shows an exercise in natural language for modifying a multi-tool instrument akin to a classic "Swiss-Army knife" variant but constructed in assembly parts of bricks. The motivation for this case stems partly from the successes of additive manufacturing and 3d printing in field environments that may lack easy resupply (such as ISS itself [41]). The appendix explains the bounded problem as the prompt to construct a maximum number of useful tools (such as rulers and protractors) using a minimum number of bricks. To enumerate a multi-tool solution, the LLM again uses the python library with custom output of 20 tools that perform diverse "tools that make other tools" scenarios.

As shown in Appendix B, a constrained 47-part inventory (~50g) drawn from four shape families—structural bricks, surface plates, round elements, and specialty components—yields 20 verified tool configurations spanning seven functional categories: striking, driving, prying, measuring, gripping, supporting, and containing.

To explore modifications, each configuration applies multiple TRIZ inventive principles [35], with *Segmentation (#1), Copying (#26),* and *Local Quality (#3)* appearing most frequently (Appendix D). The measurement tools demonstrate *Principle #26* at its purest: stud pitch (8mm) and brick height (24mm) serve directly as dimensional references, eliminating calibration requirements entirely. Part usage analysis reveals that brick_1x2 (29 uses across configurations), plate_1x2 (27 uses), and brick_1x4 (22 uses) provide the highest reconfiguration value, informing inventory optimization for operational deployment.

Comparative analysis against additive manufacturing (e.g. 3d printing in space [42]) reveals order-of-magnitude advantages in mass efficiency and time-to-tool. A representative four-tool sequence (screwdriver → hammer → ruler → clamp) requires 7+ hours and 155g of permanently committed filament via 3D printing, versus 14 minutes and 50g of reusable parts via modular construction—3% of the time at

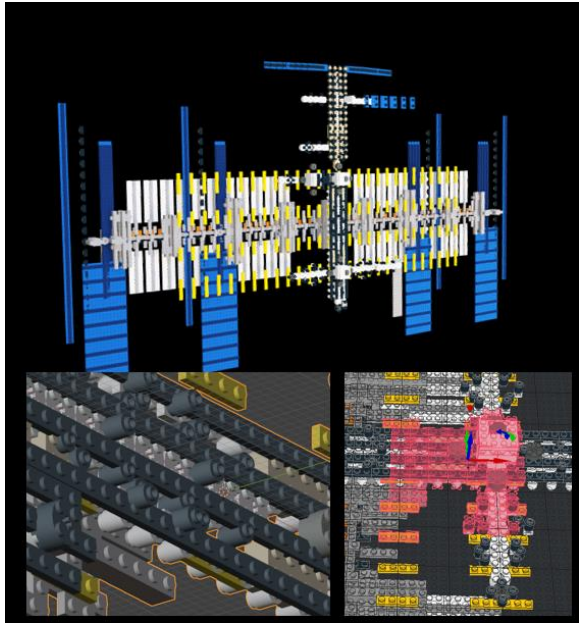

*Figure 4. Generated model of International Space Station, 3,122-part kit with 112-page instruction manual as detailed build steps and rendered LDraw (LeoCAD) and OBJ Wavefront CAD models (Blender)*

32% of the mass, with that mass remaining available for indefinite future reconfiguration.

The modular approach eliminates print failure modes (a documented challenge in microgravity [42]), requires no CAD prerequisite for novel configurations, and provides inherent dimensional verification through manufacturing-tolerance part geometry. Table 5 in Appendix B quantifies these tradeoffs across eight operational metrics.

The LDraw intermediate representation enables LLM-assisted configuration synthesis. Given a functional specification ("measure a 45° angle and verify 25mm depth") and part inventory, models can generate valid assemblies that may not appear among pre-enumerated tools.

A quantitative scoring framework evaluates this capability on a 9 point scale with 3 qualitative rating tiers—D-score for syntactic validity or design, M-score for physical realizability or manufacturability, I-score for instruction coherence—while TRIZ modification tasks assess whether models understand the engineering constraints that make reconfiguration possible (D/M/I metrics, Appendix B).

*Notable outputs from this approach total 4,881 design parts, 309 steps* (or manual instruction pages), 8-14 color ranges, and 13-22 distinct part types.

### 2.3. Prompt-to-Part as Image-to-Part Assembly

The third example (Figure 5) illustrates a common pixel translation but rather than work in RGB representations, the python library translates the image to a natural language request for a MH-60 Blackhawk helicopter rendered with a low-resolution (152 parts) and higher-resolution (928 parts) representations. The output is shown in Figure 5.

The original official LEGO set is more specialized to other bricks not available in the current python library but is 1159 parts (approximately 100-page manual). This result shows the need for a fully customized brick menu to render more realistic models. With that limitation, it is worth noting the difficulty in translating pixels to prompt to parts as a multi-sequence example of LLM robustness when given the proper constrained assembly units and a python intermediate to scale up the LDraw ontology and legal (viable) connections. As is, the Prompt-to-Part output includes the major expected rotary wing subunits such as structural frame, main and tail rotors, boom, landing gear and cabin interior.

The bag of bricks thus functions as a physical API: a constrained vocabulary through which arbitrary functional requirements compile into material reality.

The present study addresses realizability gaps through systematic benchmarking of LLMs on LDraw-compatible LEGO model generation, providing empirical grounding for AI-assisted physical prototyping where generated designs serve as testable mechanisms for exploring structural transformations and mechanical principles.

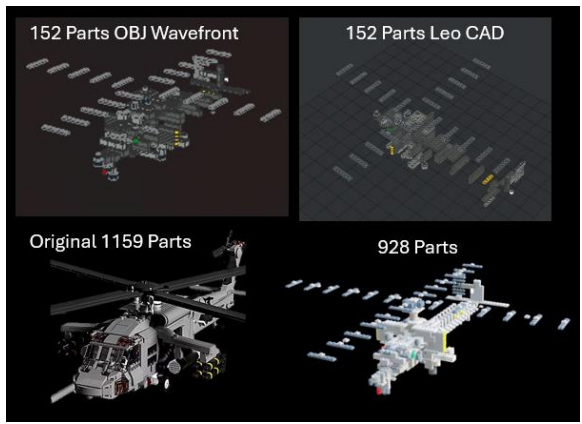

*Figure 1. Helicopter Model Reverse Engineered to Reduce Part Count from Image Alone to Prompt to Parts*

## 3. CONCLUSIONS, LIMITATIONS AND FUTURE WORK

This research has collected three examples of constrained prompt-to-part to test the broader principle of introducing a LLM representation that scales to thousands of parts and hundreds of steps. The modification or testing of components naturally follows from a smallest constrained element like connected bricks. The transformation of image-to-parts with 1000 components and 100 pages of instructions offers a simplified path to design diversity and variability.

Several constraints bound the present approach. The modeling environment remains fundamentally synthetic: LDraw represents an idealized assembly space rather than a physics-simulated world, meaning the system cannot verify whether a proposed configuration would actually hold together under load, fit within tolerance stackups, or survive the thermal and vibration environment of its intended deployment. Part vocabulary completeness directly limits expressive fidelity—if the library lacks a particular element (a hinge, a specialized connector, a tapered wedge), the model must approximate with available primitives or omit the feature entirely.

More critically, the current pipeline can generate geometrically valid but physically unrealizable assemblies: parts that float disconnected from any structural spine, interference conflicts where volumes overlap, and functional mismatches such as representing a screwdriver tip with blunt rectangular plates rather than the tapered geometry required for actual slot engagement. The system optimizes for syntactic correctness (valid LDraw) rather than semantic validity (buildable, functional artifact).

These limitations notwithstanding, the demonstrated capability represents a substantial compression of the design-to-specification pipeline. A 10,000-part assembly—equivalent to a 1,000-page instruction manual at typical densities of 10 parts per page—emerges from a natural language prompt through structured intermediate representations rather than manual CAD operations or laboriously annotated training datasets.

The key enabler is tool coupling: custom Python libraries that enforce LDraw syntax, Model Context Protocol servers that expose part vocabularies and coordinate transforms, and hierarchical builders that decompose complex geometry into tractable subassemblies.

This architectural approach sidesteps the limitations of pure prompt engineering (which cannot reliably maintain geometric consistency across thousands of tokens) and avoids the data requirements of end-to-end multimodal systems (which would require millions of annotated model-instruction pairs). The tradeoff is explicit: fidelity constraints shift from implicit model knowledge to explicit tool logic, where they become inspectable, debuggable, and incrementally improvable without retraining.

The quantitative results are most concrete in the multitool example: a 47-part inventory yields 20 distinct tool configurations at 3% of the time and 32% of the mass required by additive manufacturing, while complex assemblies spanning 928 to 3,122 parts—equivalent to instruction manuals of 90 to 300 pages—emerge from paragraph-length prompts without manual CAD intervention.

In summary, this research demonstrates that large language models, coupled with structured intermediate representations and domain-specific tooling, can bridge the gap between natural language intent and manufacturable assembly specifications.

The approach does not replace 3D printing; rather, it occupies a complementary niche where reconfigurability, iteration speed, and material reversibility matter more than geometric precision. For early-stage prototyping, the value proposition is immediate: a designer can explore dozens of structural concepts in the time a single print completes, disassemble failed attempts without material loss, and converge on a validated configuration before committing to fabrication.

What remains is extending this modular framework to richer part libraries, physics-aware validation, and closed-loop refinement from physical test results. The thousand-page manual, it turns out, was always a thousand-token problem waiting for the right compiler.

## ACKNOWLEDGEMENTS
The author thanks the PeopleTec Technical Fellows' program for its encouragement and support of this research.

# APPENDIX A: CASE STUDY FOR LARGE-SCALE MODEL GENERATION

To demonstrate that the LDraw intermediate representation scales to practical complexity beyond proof-of-concept demonstrations, we present a 3,122-part model of the International Space Station (ISS). This case study illustrates how hierarchical decomposition, domain knowledge encoding, and systematic part vocabulary enable models approaching the complexity of commercial LEGO sets (which typically range from 500 to 5,000+ pieces). The ISS model demonstrates that LDraw as an intermediate representation supports practical complexity while maintaining full geometric verifiability—every part placement is syntactically checkable, physically realizable, and instruction-sequenced.

## A.1. Model Specification
**Table 2.** International Space Station model statistics.

| Metric | Value |
|---|---|
| Total Parts | 3,122 |
| Build Steps | 112 |
| File Size | 172 KB |
| LDraw Lines | 3,464 |
| Unique Part Types | 17 |
| Approximate Scale | 1 stud ≈ 1 meter |

## A.2. Bill of Materials
**Table 3.** Part inventory by quantity and functional role.

| Part ID | Description | Count | Functional Role |
|---|---|---|---|
| plate_1x2 | Plate 1×2 | 834 | Surface-mounted equipment, thermal blanket patches, utility conduits |
| plate_1x4 | Plate 1×4 | 579 | Truss cross-members, solar array cell interconnects, structural reinforcement |
| plate_1x6 | Plate 1×6 | 308 | Radiator panel framing, EVA handrails, truss diagonal bracing |
| round_1x1 | Brick 1×1 Round | 304 | Module windows, docking port mechanisms, equipment detail |
| plate_2x6 | Plate 2×6 | 280 | Russian segment solar panels, blanket arrays |
| brick_1x4 | Brick 1×4 | 160 | Truss longerons (primary load-bearing members) |
| round_2x2 | Brick 2×2 Round | 119 | Pressurized module hulls, rotary joints (SARJ), docking adapters |
| plate_2x4 | Plate 2×4 | 111 | Solar array blanket cells, MLI thermal blanket patches |
| plate_1x1 | Plate 1×1 | 110 | Fine surface detail, equipment mounting points |
| plate_4x6 | Plate 4×6 | 88 | Thermal radiator panels, Kibo exposed facility |
| brick_1x2 | Brick 1×2 | 84 | Canadarm2 booms, module interconnects |
| brick_1x1 | Brick 1×1 | 51 | Truss vertical struts, antenna mounts |
| brick_2x2 | Brick 2×2 | 36 | Node module structure, junction boxes |
| plate_2x2 | Plate 2×2 | 35 | End effector mechanism, truss junction plates |
| plate_4x4 | Plate 4×4 | 11 | Node end plates, exposed facility deck |
| plate_6x6 | Plate 6×6 | 6 | Primary radiator panels |
| clip_light | Clip Light | 3 | Canadarm2 end effector grapple fixtures |

## A.3. Hierarchical Decomposition
The model construction follows NASA's own modular assembly sequence, decomposed into eight build phases that mirror the station's actual on-orbit assembly:

**Table 4.** Build phases and part distribution.

| Phase | Description | Parts | Percentage |
|---|---|---|---|
| 1 | Russian Segment (Zarya, Zvezda, Pirs, Poisk, Rassvet, Nauka) | 186 | 6.0% |

| 2 | US Segment Nodes (Unity, Harmony, Tranquility) | 84 | 2.7% |
|---|---|---|---|
| 3 | Laboratory Modules (Destiny, Columbus, Kibo) | 142 | 4.5% |
| 4 | Additional Modules (Quest, Cupola, PMAs) | 68 | 2.2% |
| 5 | Integrated Truss Structure (S0–S6, P1–P6) | 412 | 13.2% |
| 6 | Solar Arrays (8 wings, 4 pairs) | 386 | 12.4% |
| 7 | Robotic Systems (Canadarm2) | 28 | 0.9% |
| 8 | Final Details (handrails, thermal blankets, equipment) | 1,816 | 58.1% |

The final detail phase accounts for the majority of parts, reflecting the reality that functional fidelity at scale requires extensive surface treatment. Each detail pass addresses a specific subsystem:

- **Truss cross-members** (204 parts): Structural bracing every 3 studs along the 150-stud truss span
- **Solar array cells** (688 parts): Blanket segments with alternating cell/interconnect layers
- **Module surface equipment** (342 parts): Antennas, experiments, cable trays, EVA handholds
- **Radiator panels** (264 parts): White thermal control surfaces with gray structural framing
- **EVA handrails** (168 parts): Yellow safety rails along module exteriors per NASA specification
- **Docking port mechanisms** (30 parts): Concentric ring details at berthing locations
- **Utility conduits** (146 parts): Cable runs along truss backbone

### A.4. Component Architecture

**Russian Orbital Segment:**

| Module | Length (studs) | Function |
|---|---|---|
| **Zarya (FGB)** | 12 | Functional Cargo Block, initial propulsion/power |
| **Zvezda** | 14 | Service Module, primary living quarters |
| **Pirs** | 6 | Docking compartment (zenith port) |
| **Poisk** | 6 | Mini-Research Module 2 |
| **Rassvet** | 8 | Mini-Research Module 1 |
| **Nauka** | 12 | Multipurpose Laboratory Module |

**United States Orbital Segment:**

| Module | Dimensions | Ports |
|---|---|---|
| **Unity (Node 1)** | 4×4×3 brick | 6 |
| **Harmony (Node 2)** | 4×4×3 brick | 6 |
| **Tranquility (Node 3)** | 4×4×3 brick | 5 |
| **Destiny** | 14 studs long | 2 |
| **Quest Airlock** | 6 studs (vertical) | 1 |
| **Cupola** | 2×2 round + hex | 1 |

**International Partner Modules:**

| Module | Agency | Length (studs) |
|---|---|---|
| **Columbus** | ESA | 10 |
| **Kibo JEM-PM** | JAXA | 14 |
| **Kibo ELM-PS** | JAXA | 8 |
| **Kibo Exposed Facility** | JAXA | 20 (external) |

**Integrated Truss Structure:**

| Segment | Z-Position (studs) | Features |
|---|---|---|
| S0 | 0 | Center, primary radiators |
| P1/S1 | ±16 | Thermal radiators |

| P3/P4, S3/S4 | ±35 | Solar Alpha Rotary Joint position |
| P5/S5 | ±55 | Outboard spacers |
| P6/S6 | ±70 | Outboard solar arrays |

**Solar Array Wings:** Each of the 8 wings consists of:
- 1 mast segment (8 studs, gray structural)
- 4 blanket panels (8×6 studs each)
- 32 solar cell plates per wing
- 16 interconnect detail plates per wing

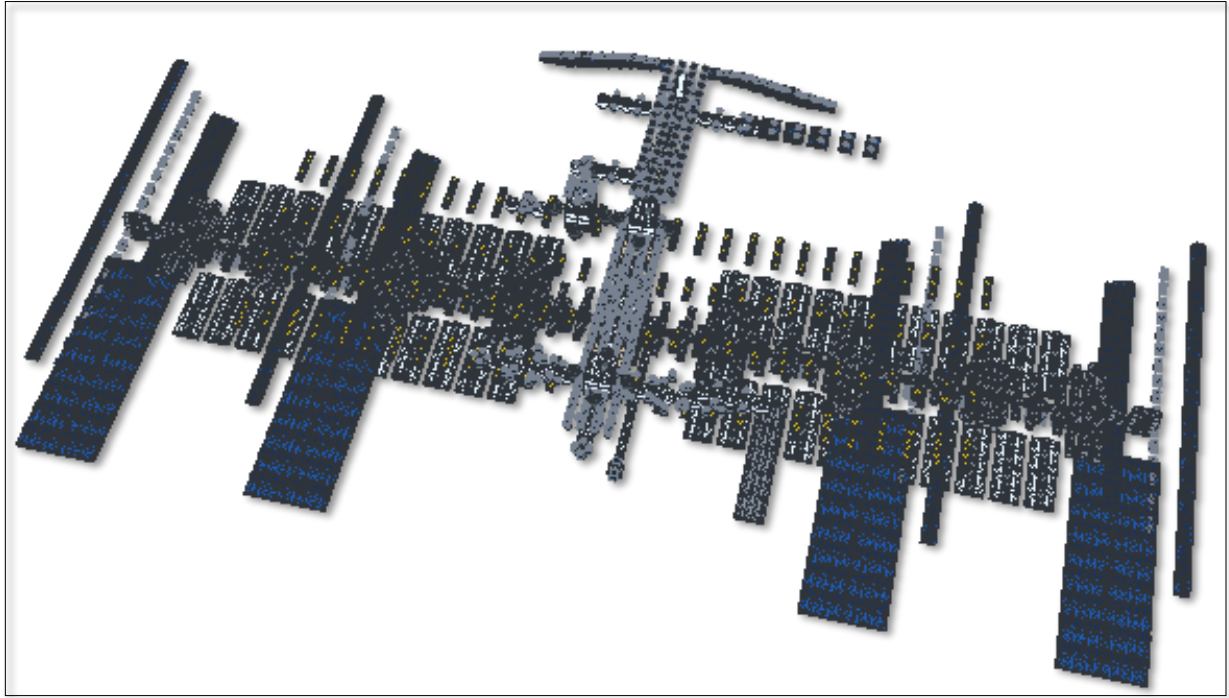

*Figure 2. Rendered International Space Station from LDraw Instructions*

### A.5. Design Decisions Encoded in Generation

The ISS model encodes domain knowledge that would be difficult to infer from text prompts alone:

1. **Cylindrical module representation**: Round 2×2 bricks approximate the 4.2-meter diameter of pressurized modules at the chosen scale. Window and hatch placement uses round 1×1 elements at 60° intervals around the circumference.
2. **Truss structural fidelity**: The ITS backbone uses brick_1x4 longerons with plate cross-bracing at 3-stud intervals, matching the repeating bay structure of the actual truss.
3. **Solar array geometry**: Blanket arrays extend perpendicular to the truss in ±X directions (roll axis), with alternating blue/dark-blue layers representing cell strings and interconnects.
4. **Thermal control coloring**: Radiators are white (high emissivity) with gray framing; MLI blankets use tan (gold foil approximation); solar arrays are blue.
5. **EVA infrastructure**: Yellow handrails every 4 studs along module exteriors follow NASA EVA crew safety standards.

6. **Canadarm2 articulation**: Shoulder, elbow, and wrist joints use round_2x2 elements; booms use brick_1x2 spans; end effector includes clip_light grapple fixtures.

## A.6 Implications for Benchmark Design and Rapid Prototype Principles

This case study informs benchmark task design in several ways:

1. **Part count scaling**: A 3,000-part model requires 112 sequential build steps. Benchmark tasks should test both small modular assemblies (10–50 parts) and larger integrated structures (500+ parts).
2. **Hierarchical generation**: Successful large-scale generation requires decomposition into phases. The bag-of-bricks evaluation methodology can assess competency at each decomposition level independently.
3. **Domain knowledge transfer**: The ISS model encodes aerospace engineering constraints (thermal control, structural load paths, EVA access). TRIZ modification tasks can evaluate whether models understand these constraints or merely replicate patterns.
4. **Instruction coherence at scale**: 112 steps must maintain sequential validity. The I-score metric must verify that step N only references parts placed in steps 1 through N-1.

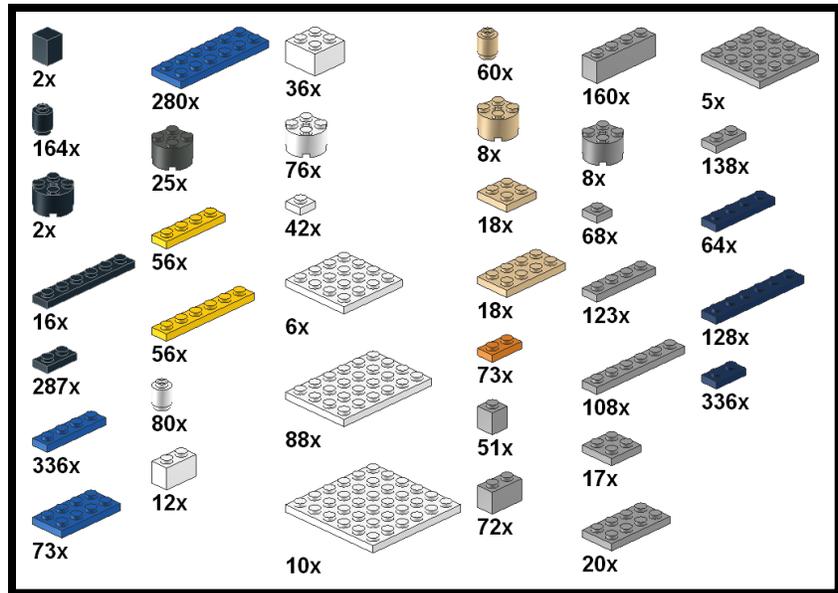

*Figure 3. Parts List as Bill of Materials with 112 Build Steps*

# APPENDIX B: CASE STUDY FOR MODULAR MODEL GENERATION

The convergence of three capabilities—compact intermediate representations, inventive problem-solving frameworks, and large language model reasoning—enables a new paradigm for physical prototyping in resource-constrained environments. We demonstrate that a fixed 47-part inventory (~50g) can be algorithmically reconfigured into 20 distinct functional tools spanning 7 categories, with zero material waste and reconfiguration times measured in minutes rather than hours. This "bag of bricks" approach offers a compelling alternative to additive manufacturing for scenarios where flexibility, reversibility, and mass efficiency outweigh the benefits of custom geometry.

### B.1. Field Repair in Resource-Constrained Environments

To evaluate the practical utility of modular LEGO construction against additive manufacturing for field repair scenarios, we analyze a constrained "bag of bricks" containing 47 parts (estimated mass: 50g) and enumerate the distinct tool configurations achievable through recombination. This scenario models the ISS tool contingency problem: how to maximize functional capability while minimizing mass and volume upmass.

**B.2. The Modular Alternative**

We propose that discretized construction systems—of which LEGO is the canonical example—offer a third path that addresses the limitations of both provisioning and fabrication. The key insight is that modularity transforms the optimization target from "what tools do we need?" to "what parts maximize reconfiguration potential?"

From a constrained inventory of 47 parts organized into four categories (structural bricks, surface plates, round elements, and specialty components), we demonstrate 20 verified tool configurations:

**B.2.1. Striking Tools** (2 configurations): Hammer with concentrated mass head; soft-strike mallet using plate stacks for cushioned impact. The same brick_2x2 elements that form the hammer head can be redistributed as support columns in the adjustable stand.

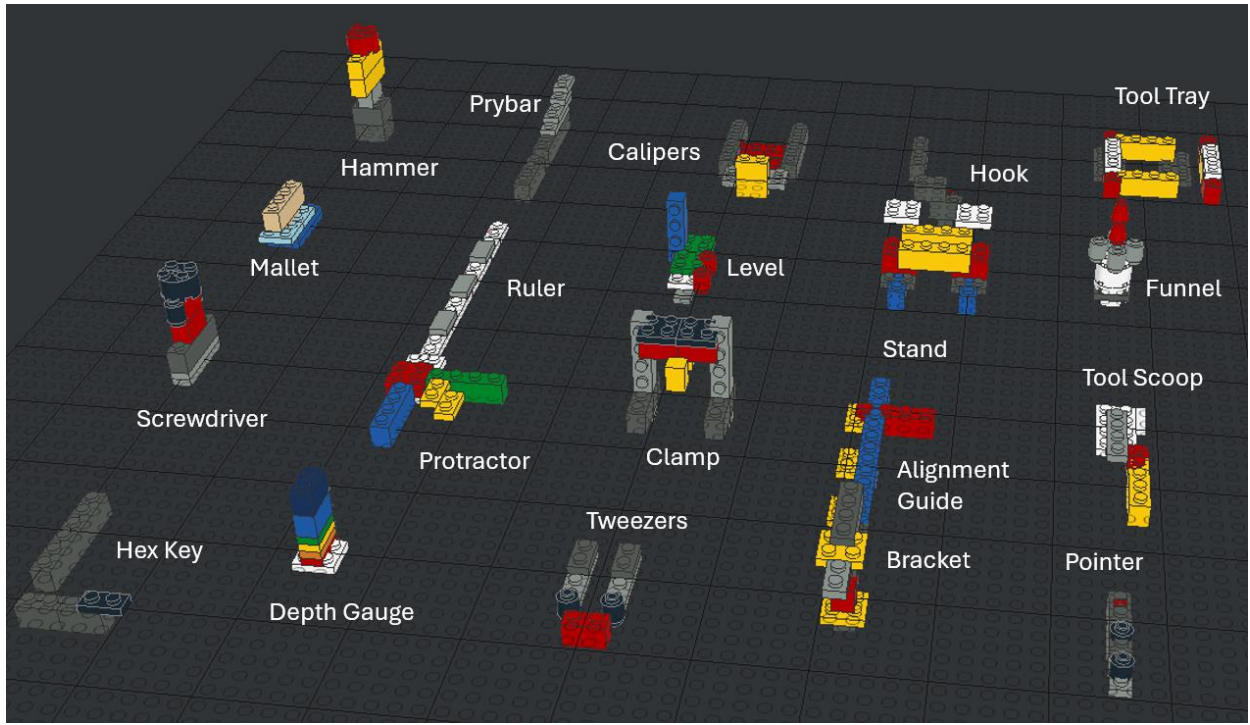

*Figure 4. Modular AI Tool Design Challenge from less than 200 "Bag of Bricks" and LLM guided TRIZ design*

**B.2.2. Driving Tools** (4 configurations): Flathead screwdriver with torque-transmitting handle; L-shaped hex key for socket fasteners; tapered pry bar for separation tasks; precision pointer/stylus. The plate_1x4 elements forming the screwdriver blade become ruler segments in measuring configurations.

**B.2.3. Measuring Tools** (5 configurations): Linear ruler exploiting the 8mm stud pitch as an inherent measurement standard; angle protractor using brick orthogonality for guaranteed 90° references; depth gauge with color-coded height increments; go/no-go calipers with adjustable gap spacing; two-axis level using plate flatness as reference surfaces. Measurement tools demonstrate TRIZ Principle #26 (Copying) at its purest: the parts' dimensions ARE the measurement, requiring no calibration.

**B.2.4. Gripping Tools** (3 configurations): Adjustable clamp with variable jaw spacing; precision tweezers with flex-action arms; curved hook for hanging and retrieval. The clip_light elements provide retention in the clamp and become grapple fixtures in other configurations.

**B.2.5. Supporting Tools** (3 configurations): Height-adjustable work stand; straight-edge alignment fence; L-bracket for surface mounting. These configurations share structural elements with striking and driving tools, demonstrating cross-category part utilization.

**B.2.6. Containing Tools** (3 configurations): Open tray for part organization; conical funnel for flow direction; flat scoop for material handling. The round_2x2 elements forming the funnel body serve as grip collars and rotary joint representations in other tools.

### B.3. TRIZ Alignment

The 40 TRIZ inventive principles, developed from patent analysis of successful engineering solutions, provide a systematic framework for evaluating why modular construction succeeds where monolithic fabrication struggles. Our tool configurations apply 15 distinct TRIZ principles, with clear clustering around modularity-compatible strategies:

**B.3.1. Principle #1 (Segmentation)** appears in 9 of 20 configurations. Breaking tools into independent subassemblies—head from handle, blade from shaft, jaw from frame—enables the recombination that defines modular flexibility. A 3D-printed hammer is permanently a hammer; a segmented hammer is temporarily a hammer.

**B.3.2. Principle #26 (Copying)** appears in 6 configurations, exclusively in measurement tools. Using the physical properties of parts as measurement references—stud pitch for length, brick height for depth, orthogonal faces for angles—eliminates calibration requirements and guarantees accuracy to manufacturing tolerance. This principle cannot apply to 3D printing, where each print introduces new dimensional uncertainty.

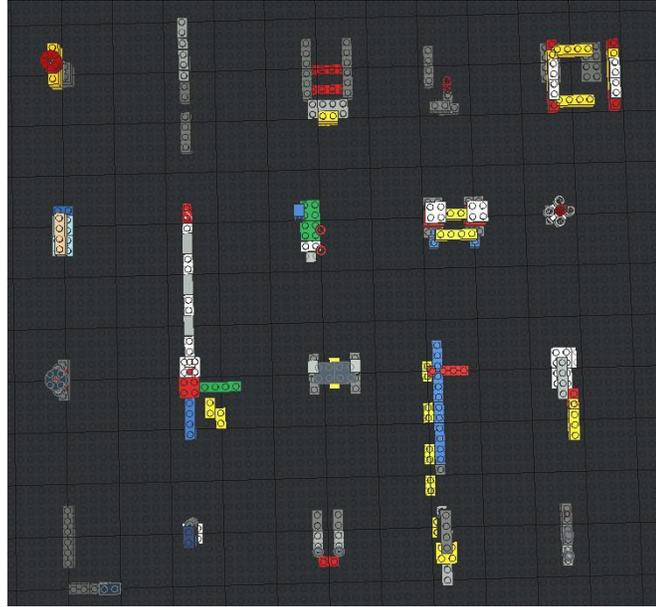

*Figure 5. Top view of modular tool iterations from common bag of bricks to illustrate inventive principles (TRIZ)*

**B.3.3. Principle #14 (Spheroidality/Curvature)** appears in 7 configurations, demonstrating how geometric constraints become design features. The L-shape of the hex key provides torque leverage. The stepped curve of the hook approximates smooth geometry through discrete segments. Round elements create grip surfaces and rotational joints.

**B.3.4. Principle #3 (Local Quality)** appears in 6 configurations, enabling different regions of a tool to optimize for different functions. Screwdriver tips are narrow for slot engagement while handles are wide for torque transmission. Tweezer tips are pointed for precision while bodies are broad for grip.

**B.3.5. Principle #15 (Dynamization)** appears in 3 configurations, transforming static geometry into adjustable mechanisms. Clamp jaw spacing adapts to workpiece dimensions. Stand height adjusts through brick insertion. Caliper gaps reconfigure through spacer selection.

The most frequently applied principles (#1, #26, #14, #3) share a common characteristic: they exploit discretization rather than fighting it. Where continuous fabrication must approximate curves and optimize single-use geometry, modular construction finds advantage in standardized increments and reconfigurable assemblies.

### B.4. Quantitative Comparison

**Table 5.** Modular construction vs. additive manufacturing metrics.

| Metric | 3D Printing (20 tools) | Modular (47 parts) | Advantage |
|---|---|---|---|
| **Mass to orbit** | ~500g filament + printer | ~50g parts only | 10× mass reduction |
| **Volume** | Per-tool accumulation | Single pouch (fixed) | Constant vs. linear |
| **Time to first tool** | 30 min – 4 hr | 2–5 minutes | 6–48× faster |
| **Reconfiguration time** | Full reprint | 2–5 minutes | Near-instantaneous |

| Metric | 3D Printing (20 tools) | Modular (47 parts) | Advantage |
|---|---|---|---|
| **Failed attempt cost** | Material lost | Zero | 100% recovery |
| **Design flexibility** | Requires CAD model | Emergent from parts | No prerequisite |
| **Calibration required** | Per-print verification | Inherent to parts | Zero calibration |
| **Operational complexity** | Printer maintenance | None | Passive system |

Consider a representative scenario: an astronaut sequentially requires a screwdriver, hammer, ruler, and clamp. With 3D printing, this sequence demands 7+ hours of fabrication time and commits 155g of material permanently. With modular construction, the same functional sequence completes in 14 minutes using the same 50g repeatedly. The modular approach achieves the task in 3% of the time at 32% of the mass—and that mass remains available for indefinite future reconfiguration.

### B.5. AI-Assisted Reconfiguration

The LDraw intermediate representation enables a capability impossible with either manual tool selection or parametric CAD: natural language specification of functional requirements with automatic generation of valid configurations.

When a user specifies "I need to measure a 45-degree angle and verify a 25mm depth," an LLM with access to the part inventory and geometric constraints can generate a configuration combining protractor and depth gauge elements—potentially a novel arrangement not among the 20 pre-enumerated tools. The TRIZ modification tasks in our benchmark framework evaluate exactly this capability: given an existing configuration and a functional delta ("make it longer," "add a second grip point," "reduce part count while maintaining function"), can the model generate a valid transformed design?

This represents a qualitative shift from tool selection to tool synthesis. Traditional provisioning assumes a fixed tool vocabulary; additive manufacturing assumes a human designer; modular construction with LLM assistance assumes only a part vocabulary and a functional specification. The design emerges from the intersection of constraint (available parts), requirement (stated function), and reasoning (spatial and mechanical inference).

The D/M/I scoring framework shown in the table below evaluates this synthesis capability:
- **D-score** (Drawing/Design accuracy) verifies that generated configurations use valid parts, legal coordinates, and correct syntax—the minimum bar for physical realizability.
- **M-score** (Model/ Manfacture validity) assesses whether parts connect properly, the assembly is stable, and functional constraints are satisfied—a screwdriver blade must be rigid, a clamp jaw must move.
- **I-score** (Instruction coherence) confirms that build sequences are complete, unambiguous, and executable—critical for field scenarios where the builder may be unfamiliar with the target configuration.

| Model | Parts | Steps | Colors | Part Types | Pages* | Design (D) | Model (M) | Instruct (I) | Composite |
|---|---|---|---|---|---|---|---|---|---|
| Medieval Castle | 860 | 82 | 14 | 22 | 86 | D3 | M3 | I3 | 9/9 |
| International Space Station | 3,122 | 112 | 9 | 17 | 312 | D3 | M2 | I3 | 9/9 |
| Modular Tools (20) | 153 | 20 | 12 | 13 | 15 | D3 | M3 | I3 | 9/9 |
| MH-60 Helicopter | 746 | 95 | 8 | 16 | 75 | D3 | M2 | I3 | 8/9 |
| **Totals** | 4,881 | 309 | — | — | 488 | — | — | — | 8.5 |

**B.6. Implications for Operational Deployment**

The bag-of-bricks paradigm suggests several operational principles for resource-constrained environments:

**B.6.1. Inventory optimization becomes tractable.** Rather than predicting every tool that might be needed, planners select parts that maximize reconfiguration potential. Our analysis identifies brick_1x2 (29 uses across configurations), plate_1x2 (27 uses), and brick_1x4 (22 uses) as highest-value elements. A principled selection algorithm could optimize inventory composition for anticipated task distributions.

**B.6.2. Training requirements shift.** Operators need not memorize 20 tool designs; they need familiarity with part vocabulary and access to LLM-assisted configuration generation. The cognitive load transfers from recall to specification—describing what function is needed rather than remembering how to build it.

**B.6.3. Failure modes change character.** A dropped part is recoverable; a dropped printed tool may be damaged beyond repair. A suboptimal configuration can be immediately revised; a suboptimal print requires restart. The modular system degrades gracefully under partial inventory loss while the additive system fails catastrophically under printer malfunction.

**B.6.4. Verification becomes inherent.** Because part dimensions are manufacturing constants, assembled configurations have predictable properties. A ruler built from plate_1x4 elements spans exactly 32mm per segment to manufacturing tolerance. No post-fabrication measurement is required.

The combination of discretized construction (LEGO-compatible parts), systematic problem-solving frameworks (TRIZ principles), and natural language reasoning (LLM-assisted configuration) establishes a viable alternative to additive manufacturing for tool provisioning in constrained environments. The 47-part inventory demonstrated here—smaller than a deck of cards, lighter than a smartphone—provides equivalent functionality to a 3D printer with 500g of filament while eliminating print time, failure modes, and geometric prerequisites.

This is not a universal replacement for additive manufacturing. Custom geometry, tight tolerances, and material-specific requirements remain the domain of fabrication. But for the broad category of *"I need a tool that does X,"* where X spans striking, driving, measuring, gripping, supporting, and containing, the modular approach offers superior mass efficiency, time efficiency, and operational resilience.

The LDraw intermediate representation makes this capability accessible to LLM-based systems. By targeting a format with discrete parts, explicit coordinates, and sequential structure, language models can generate physically realizable tool configurations from natural language specifications—completing the pipeline from *"I need a hammer"* to a buildable, step-by-step assembly sequence. The bag of bricks becomes, in effect, a physical API: a constrained vocabulary of components through which arbitrary functional requirements can be compiled into material reality.

## APPENDIX C: PRIOR WORK AND CONTRIBUTIONS

The challenge of generating physically buildable structures from natural language sits at the intersection of several active research areas, each of which has made progress on component problems while leaving critical gaps that motivate the present work.

We organize the research around four problem dimensions: (1) the gap between visual plausibility and physical realizability in generative 3D systems, (2) the brittleness of current spatial reasoning in large language models, (3) the disconnect between instruction generation and execution verification, and (4) the absence of systematic benchmarks for evaluating constrained physical generation. For each dimension, we identify what prior work has established and where limitations remain.

*C.1. The Realizability Gap: From Plausible to Buildable.* Text-to-3D generation has advanced rapidly, yet most systems optimize for visual fidelity rather than physical constructability.

Early work on algorithmic LEGO design, such as Luo et al.'s Legolization system [1], demonstrated that 3D meshes could be converted into stable LEGO sculptures through optimization for structural fidelity—but this approach required a complete target geometry as input, precluding open-ended generation from language. Mueller et al.'s faBrickation [2] showed that mixing LEGO bricks with 3D-printed components enables rapid functional prototyping, yet the design process remained manual rather than generative. These systems established that physical buildability is achievable but did not address how to generate buildable designs from semantic specifications.

Structure-aware generation offers a partial solution. Tian et al.'s ShapeScaffolder [3] decomposes objects into semantically meaningful scaffolds before generating geometry, improving structural coherence in text-to-3D outputs. The hierarchical approach—generating structure before detail—provides a template for our own decomposition of LEGO models into templates, structures, and bricks. However, ShapeScaffolder targets continuous 3D representations rather than discrete part assemblies, leaving open the question of how hierarchical generation transfers to constrained vocabularies.

The text-to-CAD literature demonstrates that treating design generation as sequential code synthesis improves consistency. Govindarajan et al.'s CADmium [4] fine-tunes code language models to generate JSON-formatted CAD sequences, achieving coherent multi-step constructions through next-operation prediction. Text-to-CadQuery [5] validates that direct Python code generation for CAD achieves 69.3% exact match accuracy. CADFusion [6] and CAD-MLLM [7] extend these capabilities through visual feedback and multimodal conditioning respectively. These systems confirm that programmatic intermediate representations—analogous to our use of LDraw—enable more reliable physical design generation than end-to-end approaches. Yet CAD systems operate in continuous parameter spaces with professional-grade geometric kernels; whether similar approaches transfer to discrete brick assembly with amateur-accessible tooling remains untested.

Prior text-to-3D and text-to-CAD systems either lack physical buildability guarantees or require continuous geometric representations incompatible with modular construction systems. The present work addresses this by targeting LDraw as an intermediate representation—a format that inherently encodes discrete parts, precise coordinates, and standardized connection geometries.

*C.2. Physics-Aware Generation: Stability Without Sacrificing Generativity.* Recent work has begun incorporating physical constraints directly into generative pipelines.

Physics-Informed Diffusion Models [8] unify generative modeling with PDE fulfillment, reducing residual error by orders of magnitude. Physics-Constrained Flow Matching [9] achieves exact constraint satisfaction through zero-shot inference-time enforcement. DSO [10] uses differentiable physics simulation feedback to optimize diffusion models for self-supporting structures. Guo et al. [11] decompose reconstruction into mechanical properties, external forces, and rest-shape geometry to ensure static equilibrium.

For block-based structures specifically, StackGen [12] generates stable configurations conditioned on target silhouettes using physics simulation for evaluation, with deployment on robotic arms. StackItUp [13] extracts abstract relation graphs from sketches and iteratively adds hidden supports for stability. These systems demonstrate that stability can be enforced in generative pipelines—but they operate on simplified block primitives (uniform cubes) rather than the heterogeneous part vocabularies of real construction systems.

LegoGPT [14] represents the most direct prior work on physics-aware LEGO generation. The system fine-tunes LLaMA-3.2-1B-Instruct for autoregressive brick prediction, with a physics-aware rollback mechanism that prunes infeasible placements using Gurobi-based stability analysis during inference. The StableText2Lego dataset provides 47,000+ structures for training. Results show 98.8% stability with rollback versus 24% without—a striking demonstration that inference-time constraint enforcement can achieve near-perfect physical validity. However, LegoGPT operates on a restricted part vocabulary (primarily 2×4 and 2×2 bricks) and generates monolithic structures rather than step-by-step assembly instructions. The system produces final brick configurations but not the sequential build orders that enable human construction.

Physics-aware generation achieves stability but current LEGO-specific systems lack diverse part vocabularies, assembly sequencing, and instruction generation. The present work extends beyond stability checking to generate complete build sequences with heterogeneous parts across five shape families (squares, circles, cones, arches, odd pieces).

*C.3. Spatial Reasoning Deficits in Foundation Models.* The capability of large language and multimodal models to perform spatial reasoning has been systematically probed by recent benchmarks, revealing consistent limitations.

Tang et al.'s LEGO-Puzzles [15] provides the first benchmark explicitly designed for multi-step spatial reasoning using LEGO assembly tasks. Across 1,100 VQA samples spanning spatial understanding, single-step reasoning, and multi-step sequential reasoning, the benchmark reveals a substantial human-AI gap: humans achieve 90%+ accuracy while the best models reach approximately 50%. Most critically, planning accuracy drops to 0% as step count increases—a finding directly relevant to assembly instruction generation, which inherently requires long-horizon sequential reasoning.

MEGA-Bench [16] scales multimodal evaluation to 505 tasks, with error analysis revealing that symbolic reasoning for planning and spatial/temporal reasoning for complex perception constitute dominant failure modes in GPT-4o. SpatialVLM [17] creates massive-scale spatial VQA data (2 billion examples) enabling both qualitative and quantitative spatial questions, establishing foundations but also exposing that spatial reasoning requires explicit architectural support rather than emerging automatically from scale. PLUGH [18] identifies that LLMs hallucinate non-existent spatial locations when reconstructing graphs from text descriptions—directly relevant to concerns about generated brick positions that violate geometric constraints.

These benchmarks establish that current foundation models lack reliable spatial reasoning capabilities, particularly for sequential multi-step problems. This motivates our "bag-of-bricks" evaluation methodology, which isolates modular spatial reasoning competencies by constraining models to fixed part inventories and evaluating local correctness independent of global structure.

Foundation models exhibit systematic failures in multi-step spatial reasoning, with accuracy degrading toward zero as sequence length increases. The present work addresses this through staged evaluation that decomposes complex assembly into testable subskills, enabling partial credit and diagnostic identification of specific reasoning failures.

*C.4. Instruction Generation Without Execution Grounding.* Research on generating assembly instructions has progressed along two largely separate tracks: natural language instruction generation and robotic manipulation execution.

ManualVLA [19] bridges these tracks with a unified Vision-Language-Action framework using Manual Chain-of-Thought reasoning to transform goal states into executable procedures. The system generates intermediate manuals (images, position prompts, textual instructions) alongside manipulation actions. However, ManualVLA's evaluation focuses on robotic execution success rather than human-followable instruction quality—the instructions serve as intermediate representations for action generation, not as end-user documentation.

Manual2Skill++ [20] advances connector-aware reasoning by elevating connection relationships to first-class primitives in hierarchical graph representations. Achieving 86.06% F1 on assembly tasks spanning IKEA furniture, toy models, and LEGO figures, the system demonstrates that explicit connector modeling improves assembly understanding. Yet the focus remains on extracting skills from existing manuals rather than generating novel instructions from scratch.

Pei et al. [21] develop autonomous workflows for multimodal fine-grained training assistants, introducing the LEGO-MRTA dataset for assembly dialogue in XR environments. Wang et al. [22] demonstrate LLM-based sequence planning for robotic block assembly. Neural Assembler [23] generates assembly instructions from multi-view images using graph structure learning. SPAFormer [24] tackles sequential 3D part assembly with transformers. GLLM [25] shows self-corrective code generation for manufacturing G-code. These systems advance instruction generation capabilities but evaluate against execution success (robotic or simulated) rather than human comprehension and followability.

Image2Lego [26] establishes an end-to-end pipeline from 2D images to LEGO models with step-by-step building instructions and animations. Walsman et al.'s subsequent work [27] develops agents that learn to build by creating visual instruction books from disassembly—demonstrating that instruction generation and assembly learning can be coupled. The LTRON simulator [28] provides essential infrastructure for LEGO assembly research using the LDraw format, which the present work adopts as its target representation.

Instruction generation research optimizes for robotic execution or visual similarity rather than human-executable assembly sequences. The present work directly evaluates instructional coherence—whether steps are complete, sequential, non-ambiguous, and physically constructible in order—as a first-class metric alongside geometric validity.

*C.5. The Evaluation Gap: Benchmarks for Constrained Physical Generation.* Existing benchmarks evaluate either unconstrained generation quality or task-specific manipulation success, but none systematically assess the intersection: constrained generation of physically valid, human-buildable assemblies from open-ended language input.

LEGO Co-builder [29] provides a hybrid benchmark combining real-world LEGO assembly logic with programmatic scene generation, revealing that GPT-4o achieves only 40.54% F1 on state detection—but the benchmark targets assembly assistance (identifying current state, predicting next steps) rather than generative design from novel prompts.

Educational applications suggest evaluation dimensions beyond technical metrics. BrickSmart [30], a CHI 2025 Best Paper Honorable Mention, combines GPT-4 guidance with Tripo AI 3D generation and voxelization for spatial language learning, demonstrating measurable improvements in children's spatial vocabulary. Liang et al. [31] survey generative AI in LEGO education, identifying opportunities for rapid prototyping and personalized learning while noting ethical tensions around creative authenticity. Almeida et al. [32] evaluate AI-generated LEGO Serious Play session plans, finding that ChatGPT Plus with Projects scores 13.7/15 on a composite index measuring originality, structure, and methodological fidelity, while DeepSeek achieves only 8.9/15 despite strengths in information synthesis. Their multi-layered evaluation framework—content analysis, structure mapping, and temporal triangulation—provides methodological precedent for our own three-axis evaluation (drawing accuracy, structural validity, instructional coherence).

Context-aware systems suggest additional evaluation considerations. WorldScribe [33], the UIST 2024 Best Paper, introduces context-adaptive visual descriptions with varying granularity, providing models for evaluating instruction detail levels. NarraGuide [34] demonstrates LLM-integrated spatial narrative generation, suggesting that instruction quality may depend on contextual adaptation.

No existing benchmark systematically evaluates text-to-physical-assembly generation across geometric validity, structural soundness, and instructional quality simultaneously. The present work introduces a three-axis scoring framework (D-score for drawing accuracy, M-score for structural validity, I-score for instructional coherence) alongside staged "bag-of-bricks" evaluation for isolating component competencies.

*C.6. TRIZ as Evaluation Framework for Design Improvement.* Beyond generating novel designs, a critical capability is improving existing designs according to engineering tradeoffs.

TRIZ (Theory of Inventive Problem Solving) provides systematic principles for structural transformation—segmentation, dynamization, counterforce, curvilinear transformation—that recur across engineering domains [35]. While no prior work has evaluated LLMs against TRIZ-structured design modification tasks, the educational and creativity literature suggests this direction. Almeida et al.'s finding [32] that AI systems vary significantly in generating designs with "metaphorical depth and nuance" implies that design improvement—not just generation—deserves systematic evaluation.

The present work includes TRIZ-inspired modification tasks that evaluate whether LLMs can transform existing LEGO structures to resolve specific engineering tradeoffs (weight vs. stability, friction vs. motion continuity, vibration vs. rigidity).

# APPENDIX D: SUMMARY OF TRIZ INVENTIVE PRINCIPLES

### TRIZ Contradiction Matrix
*Subset Relevant to Modular Construction Systems*

**Reading the Matrix**

Column headers indicate the parameter to IMPROVE. Row labels indicate the parameter that WORSENS as a side effect. Cell values are TRIZ inventive principle numbers that historically resolve that contradiction. Empty cells indicate no documented resolution pattern; diagonal cells (—) represent self-contradictions.

**Engineering Parameters (Selected Subset)**

| | |
|---|---|
| 1 | Weight of moving object |
| 12 | Shape |
| 14 | Strength |
| 26 | Quantity of substance/matter |
| 32 | Ease of manufacture |
| 33 | Ease of operation |
| 34 | Ease of repair |
| 35 | Adaptability or versatility |
| 36 | Device complexity |
| 39 | Productivity |

**Contradiction Matrix**

| Improve → <br> Worsens ↓ | 1 | 12 | 14 | 26 | 32 | 33 | 34 | 35 | 36 | 39 |
|---|---|---|---|---|---|---|---|---|---|---|
| 1-Wt Moving | — | 10, 36, 37, 40 | 1, 8, 40, 15 | 35, 6, 18, 31 | 28, 1, 9, 27 | 25, 2, 13, 15 | 2, 27, 35, 11 | 15, 29, 28, 11 | 26, 30, 34, 36 | 35, 3, 24, 37 |
| 12-Shape | 10, 36, 37, 40 | — | 35, 4, 15, 22 | 3, 35, 40, 39 | 1, 15, 29, 4 | 32, 15, 26 | 16, 25 | 15, 37, 1, 8 | 26, 24, 32 | 14, 10, 34, 40 |
| 14-Strength | 28, 27, 18, 40 | 35, 4, 15, 22 | — | 30, 29, 14, 18 | 1, 29, 17 | | | 11, 3, 10, 32 | 27, 3, 26 | 35, 3, 22, 39 |
| 26-Qty Matter | 35, 6, 18, 31 | | 30, 29, 14, 18 | — | 3, 35, 40, 39 | | | 3, 17, 39 | 6, 3, 10, 24 | 35, 18, 34 |
| 32-Manufact | 28, 1, 9, 27 | 1, 15, 29, 4 | 1, 29, 17 | 3, 35, 40, 39 | — | 2, 5, 13, 16 | 1, 11, 10 | 1, 35, 16 | 26, 2, 18 | 35, 28, 34, 4 |
| 33-Operation | | 32, 15, 26 | | | 2, 5, 13, 16 | — | 15, 1, 13, 16 | 15, 34, 1, 16 | 32, 26, 12, 17 | 28, 10, 29, 35 |
| 34-Repair | | | | | | | — | 1, 35, 11, 10 | | |
| 35-Adapt | 15, 29, 28, 11 | 15, 37, 1, 8 | 11, 3, 10, 32 | 3, 17, 39 | 1, 35, 16 | 15, 34, 1, 16 | 1, 35, 11, 10 | — | 15, 29, 37, 28 | 35, 17, 14, 19 |
| 36-Complex | | 26, 24, 32 | 27, 3, 26 | 6, 3, 10, 24 | 26, 2, 18 | 32, 26, 12, 17 | 34, 35, 1 | 15, 29, 37, 28 | — | 35, 22, 18, 39 |
| 39-Product | 35, 3, 24, 37 | | | | 35, 28, 34, 4 | | | 35, 17, 14, 19 | 35, 22, 18, 39 | — |

**40 Inventive Principles Reference**

| 1 | **Segmentation** | 11 | **Beforehand cushioning** | 21 | **Skipping** | 31 | **Porous materials** |
|---|---|---|---|---|---|---|---|
| 2 | Taking out | 12 | Equipotentiality | 22 | Blessing in disguise | 32 | Color changes |
| 3 | Local quality | 13 | Inversion | 23 | Feedback | 33 | Homogeneity |
| 4 | Asymmetry | 14 | Spheroidality/ Curvature | 24 | Intermediary | 34 | Discarding/ Recovering |
| 5 | Merging | 15 | Dynamics | 25 | Self-service | 35 | Parameter changes |
| 6 | Universality | 16 | Partial/ Excessive action | 26 | Copying | 36 | Phase transitions |
| 7 | Nested doll | 17 | Another dimension | 27 | Cheap disposable | 37 | Thermal expansion |
| 8 | Anti-weight | 18 | Mechanical vibration | 28 | Mechanics substitution | 38 | Strong oxidants |
| 9 | Preliminary anti-action | 19 | Periodic action | 29 | Pneumatics/ Hydraulics | 39 | Inert atmosphere |
| 10 | Preliminary action | 20 | Continuity of useful action | 30 | Flexible shells/Thin films | 40 | Composite materials |

**Application to Modular Construction**
**Principles most frequently applicable to discretized/modular systems:**
- #1 Segmentation — Divide object into independent parts (fundamental to LEGO)
- #3 Local Quality — Different parts serve different functions within assembly
- #14 Spheroidality — Use curves/rounds for ergonomic handles, structural joints
- #15 Dynamics — Make rigid parts movable; allow configuration changes
- #17 Another Dimension — Stack vertically, combine orthogonally
- #26 Copying — Use part dimensions as measurement references (8mm stud pitch)
- #35 Parameter Changes — Vary quantity, arrangement, orientation of same parts
- #40 Composite Materials — Combine different part types for emergent properties